\def\year{2020}\relax
\documentclass[letterpaper]{article} 
\usepackage{aaai20}  
\usepackage{times}  
\usepackage{helvet} 
\usepackage{courier}  
\usepackage[hyphens]{url}  
\usepackage{graphicx} 
\usepackage{graphicx}  
\usepackage{bm}
\usepackage{amssymb}
\usepackage{mathrsfs}
\usepackage{xcolor}
\usepackage{amsmath}
\usepackage{multirow}
\newcommand{\citet}[1]{\citeauthor{#1} \shortcite{#1}}
\newcommand{\citep}{\cite}
\newcommand{\citealp}[1]{\citeauthor{#1} \citeyear{#1}}
\usepackage{url}

\title{Learning Cross-Modal Context Graph for Visual Grounding}

\author{Yongfei Liu\textsuperscript{\rm 1}\thanks{Both authors contributed equally to the work. This work was supported by Shanghai NSF Grant (No. 18ZR1425100) and NSFC Grant (No. 61703195). The research of 3rd author is supported by the Discovery Grant of Natural Sciences and Engineering Research Council of Canada.}\quad Bo Wan\textsuperscript{\rm 1}\footnotemark[1] \quad Xiaodan Zhu\textsuperscript{\rm 2} \quad Xuming He\textsuperscript{\rm 1}
\\ \Large \textsuperscript{\rm 1} ShanghaiTech University\quad \textsuperscript{\rm 2} Queen’s University \\
\{liuyf3, wanbo, hexm\}@shanghaitech.edu.cn\quad xiaodan.zhu@queensu.ca}
 
\begin{document}

\maketitle

\begin{abstract}
        Visual grounding is a ubiquitous building block in many vision-language tasks and yet remains challenging due to large variations in visual and linguistic features of grounding entities, strong context effect and the resulting semantic ambiguities. Prior works typically focus on learning representations of individual phrases with limited context information. To address their limitations, this paper proposes a language-guided graph representation to capture the global context of grounding entities and their relations, and develop a cross-modal graph matching strategy for the multiple-phrase visual grounding task. In particular, we introduce a modular graph neural network to compute context-aware representations of phrases and object proposals respectively via message propagation, followed by a graph-based matching module to generate globally consistent localization of grounding phrases. We train the entire graph neural network jointly in a two-stage strategy and evaluate it on the Flickr30K Entities benchmark. Extensive experiments show that our method outperforms the prior state of the arts by a sizable margin, evidencing the efficacy of our grounding framework. Code is available at \url{https://github.com/youngfly11/LCMCG-PyTorch}.
\end{abstract}

\section{Introduction}

Integrating visual scene and natural language understanding is a fundamental problem toward achieving human-level artificial intelligence, and has attracted much attention due to rapid advances in computer vision and natural language processing~\cite{aditya2019trends}. A key step in bridging vision and language is to build a detailed correspondence between a visual scene and its related language descriptions. In particular, the task of grounding phrase descriptions into their corresponding image has become an ubiquitous building block in many vision-language applications, such as image retrieval~\cite{image_retrieval1,image_retrieval2}, image captioning~\cite{li2017msdn,feng2019unsupervised}, visual question answering~\cite{NIPS2018_8031,Cadene_2019_CVPR} and visual dialogue~\cite{abhishek2017vd,Kottur_2018_ECCV}. 

General visual grounding typically attempts to localize object regions that correspond to \textit{multiple} noun phrases in image descriptions. 
Despite significant progress in solving  vision~\cite{ren2015faster,zhang2017visual} or language~\cite{Peters2018,devlin2018bert} tasks, it remains challenging to establish such cross-modal correspondence between objects and phrases, mainly because of large variations in object appearances and phrase descriptions, strong context dependency among these grounding entities, and the resulting semantic ambiguities in their representations~\cite{flickrentities,plummer2018conditional}. 

Many existing works on visual grounding tackle the problem by localizing each noun phrase independently via phrase-object matching~\cite{flickrentities,plummer2018conditional,yu2018rethinking,GroundingR}. However, such grounding strategy tends to ignore visual and linguistic context, thus leading to matching ambiguity or errors for complex scenes. 
Only a few grounding approaches take into account context information~\cite{SeqGROUND,chen2017query} or phrase relationship~\cite{wang2016structured,plummerPLCLC2017} when representing visual or phrase entities. While they partially alleviate the problem of grounding ambiguity, their context or relation representations have several limitations for capturing global structures in language descriptions and visual scenes. First, for language context, they typically rely on chain-structured LSTMs defined on description sentences, which have difficulty in encoding long-range dependencies among phrases. In addition, most methods simply employ off-the-shelf object detectors to generate object candidates for cross-modal matching. However, it is inefficient to encode visual context for those objects due to a high ratio of false positives in such object proposal pools. Furthermore, when incorporating phrase relations, these methods often adopt a stage-wise strategy that learns representations of noun phrases and their relationship separately, which is sub-optimal for the overall grounding task.

In this work, we propose a novel cross-modal graph network to address the aforementioned limitations for multiple-phrase visual grounding. Our main idea is to exploit the language description to build effective global context representations for all the grounding entities and their relations, which enables us to generate a selective set of high-quality object proposals from an image and to develop a context-aware cross-modal matching strategy. To achieve this, we design a modular graph neural network consisting of four main modules: a backbone network for extracting basic language and visual features, a phrase graph network for encoding phrases in the sentence description, a visual object graph network for computing object proposal features and a graph similarity network for global matching between phrases and object proposals.    

Specifically, given an image and its textual description, we first use the \textit{backbone network} to compute the language embedding for the description, and to generate an initial set of object proposals. 
To incorporate language context, we construct a language scene graph from the description~(e.g.,~\citealp{schuster2015generating};~\citealp{wang-etal-2018-scene}) in which the nodes are noun phrases, and the edges encode relationships between phrases.
Our second module, \textit{phrase graph network}, is defined on this language scene graph and computes a context-aware phrase representation through message propagation on the phrase graph. 
We then use the phrase graph as a guidance to build a visual scene graph, in which the nodes are object proposals relevant to our phrases, and the edges encode the same type of relations as in the phrase graph between object proposals. The third network module, \textit{visual object graph network}, is defined on this derived graph and generates a context-aware object representation via message propagation. 
Finally, we introduce a \textit{graph similarity network} to predict the global matching of those two graph representations, taking into account similarities between both graph nodes and relation edges.  

We adopt a two-stage strategy in our model learning, of which the first stage learns the phrase graph network and visual object features while the second stage trains the entire deep network jointly. 
We validate our approach by extensive experiments on the public benchmark Flickr30K Entities~\cite{flickrentities}, and our method outperforms the prior state of the art by a sizable margin. To better understand our method, we also provide the detailed ablative study of our context graph network.

The main contributions of our work are three-folds:
\begin{itemize}
	\item We propose a language-guided graph representation, capable of encoding global contexts of phrases and visual objects, and a globally-optimized graph matching strategy for visual grounding.
	\item We develop a modular graph neural network to implement the graph-based visual grounding, and a two-stage learning strategy to train the entire model jointly. 
	\item Our approach achieves new state-of-the-art performance on the Flickr30K Entities benchmark.
\end{itemize}

\section{Related Works}

\subsubsection{Visual Grounding:}
In general, visual grounding aims to localize object regions in an image corresponding to multiple noun phrases from a sentence that describes the underlying scene. 
Rohrbach et al.~\shortcite{GroundingR} proposed an attention mechanism to attend to relevant object proposals for a given phrase and designed a loss for phrase reconstruction.
Plummer et al.~\shortcite{plummer2018conditional} presented an approach to jointly learn multiple text-conditioned embedding in a single end-to-end network. In DDPN~\cite{yu2018rethinking}, they learned a diversified and discriminate proposal network to generate higher quality object candidates. Those methods grounded each phrase independently, ignoring  the context information in image and language. Only a few approaches attempted to solve visual grounding by utilizing context cues. 
Chen et al.~\shortcite{chen2017query} designed an additional reward by incorporating context phrases and train the whole network by reinforcement learning.
Dongan et al.~\shortcite{SeqGROUND} took context into account by adopting chain-structured LSTMs network to encode context cues in language and image respectively.
In our work, we aim to build cross-modal graph networks under the guidance of language structure to learn global context representation for grounding entities and object candidates.

\subsubsection{Referring Expression:} Referring expression comprehension is closely related to visual grounding task, which attempts to localize expressions corresponding to image regions. Unlike visual grounding, those expressions are typically region-level descriptions without specifying grounding entities.
Nagaraja et al.~\shortcite{nagaraja2016modeling} proposed to utilize LSTMs to encode visual and linguistic context information jointly for referring expression. Yu et al.~\shortcite{yu2018mattnet} developed modular attention network, which utilized language-based attention and visual attention to localize the relevant regions. Wang et al.~\shortcite{peng2019neighbor} applied self-attention mechanism on sentences and built a directed graph over neighbour objects to model their relationships.
All the above-mentioned methods fail to explore the structure of the expression explicitly. Our focus is to exploit the language structure to extract cross-modal context-aware representations.

\subsubsection{Structured Prediction:} Structured prediction is a framework to solve the problems whose output variables are mutually dependent or constrained. 
Justin et al.~\shortcite{image_retrieval1} proposed the task of scene graph grounding to retrieve images, and formulated the problem as structured prediction by taking into account both object and relationship matching. 
To explore the semantic relations in visual grounding task, Wang et al.~\shortcite{wang2016structured} tried to introduce a relational constraint between phrases, but limited their relations to possessive pronouns only. Plummer et al.~\shortcite{plummerPLCLC2017} extended the relations to attributes, verbs, prepositions and pronouns, and performed global inference during test stage. 
We extend these methods by exploiting the language structure to get context-aware cross-modal representations and learn the matching between grounding entities and their relations jointly.


\begin{figure*}
	\centering
	\includegraphics[width=0.8\linewidth]{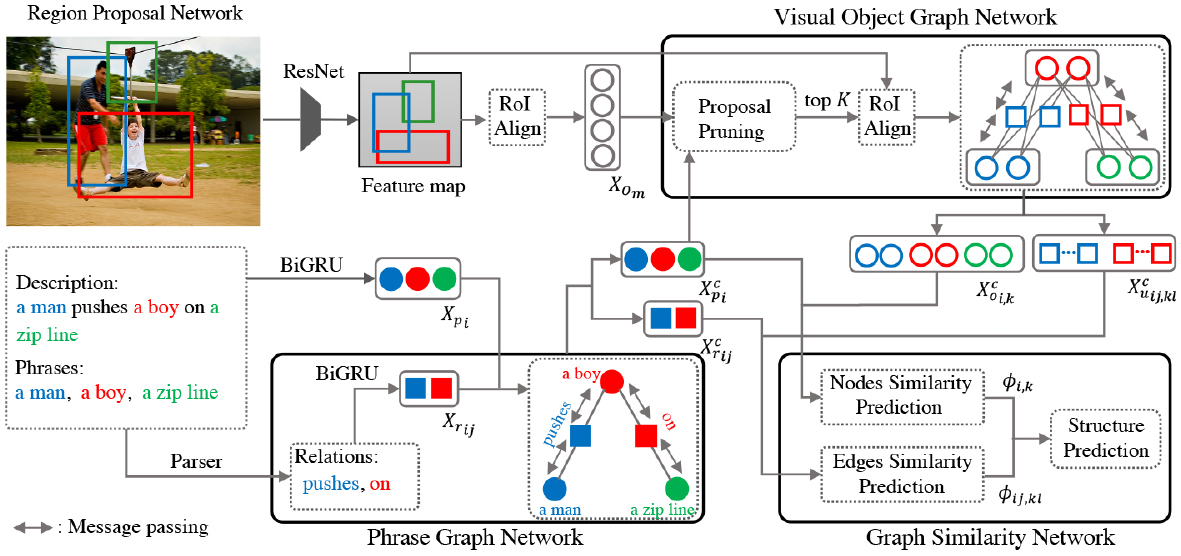}
	\caption{\small{\textbf{Model Overview}: There are four modules in our network, the \textbf{Backbone Network} extracts basic linguistic and visual features; the \textbf{Phrase Graph Network} is defined on the a parsed language scene graph to refine language representations; the \textbf{Visual Object Graph Network} is defined on a visual scene graph which is constructed under the guidance of the phrase graph to refine visual object feature; finally a \textbf{Graph Similarity Network} predicts the global matching of those two graph representations. \textbf{\textit{Solid circles}} denote noun phrase features while \textbf{\textit{solid squares}} represent relation phrase features. \textbf{\textit{Hollow circles and squares}} denote visual object and relation features respectively.
	} }
	\label{fig:model}
\end{figure*}

\section{Problem Setting and Overview}
The task of general visual grounding aims to localize a set of object regions in an image, each corresponding to a noun phrase in a sentence description of the image. 
Formally, given an image $I$ and a description $Q$, we denote a set of noun phrases for grounding as $\mathcal{P}=\{p_i\}_{i=1}^N$ and their corresponding locations as $\mathcal{B}=\{b_i\}_{i=1}^N$ where $b_i\in \mathbb{R}^4$ is the bounding box parameters. Our goal is to predict the set $\mathcal{B}$ for a given set $\mathcal{P}$ from the input $I$ and $Q$.

To this end, we adopt a hypothesize-and-match strategy that first generates a set of object proposals $\mathcal{O}=\{o_m\}_{m=1}^M$ and then formulates the grounding task as a matching problem, in which we seek to establish a cross-modal correspondence between the phrase set $\mathcal{P}$ and the object proposal set $\mathcal{O}$. This matching task, nevertheless, is challenging due to large variations in visual and linguistic features, strong context dependency among the grounding entities and the resulting semantic ambiguities in pairwise matching. 

To tackle those issues, we propose a language-guided approach 
motivated by the following three key observations: 
First, language prior can be used to generate a graph representation of noun phrases and their relations, which captures the global context dependency more effectively than chain-structured models. 
In addition, the object proposals generated by detectors typically have a high ratio of false positives, and hence it is difficult to encode visual context for each object. We can exploit language structure to guide proposal pruning and build a better context-aware visual representation. 
Finally, the derived phrase graph structure also includes the phrase relations, which provide additional constraints in the matching for mitigating ambiguities.

We instantiate these ideas by designing a cross-modal graph network for the visual grounding task, which consists of four main modules: 
a) a \textit{backbone network} that extracts basic linguistic and visual features;
b) a \textit{phrase graph network} defined on a language scene graph built from the description to compute the context-aware phrase representations; c) a \textit{visual graph network} defined on a visual scene graph of object proposals constructed under the guidance of the phrase graph, and encodes context cues for the object representations via message propagation;
and d) a \textit{graph similarity network} that predicts a global matching of the two graph representations. The overall model is shown in Fig.~\ref{fig:model} and we will describe the details of each module in the following section.

\section{Cross-modal Graph Network}
We now introduce our cross-modal graph matching strategy, including the model design of four network modules and the overall inference pipeline, followed by our two-stage model training procedure.  

\subsection{Backbone Network}\label{backbone}
Our first network module is a backbone network that takes as input the image $I$ and description $Q$, and generates corresponding visual and linguistic features. The backbone network consists of two sub-networks: a convolutional network for generating object proposals and a recurrent network for encoding phrases. 

Specifically, we adopt the ResNet-101\cite{he2016deep} as our convolutional network to generate feature map $\mathbf{\Gamma}$ with channel dimension of $D_0$. We then apply a Region Proposal Network (RPN)~\cite{ren2015faster} to generate an initial set of object proposals $\mathcal{O}=\{o_m\}_{m=1}^M$, where $o_m\in \mathbb{R}^4$ denotes object location~(i.e. bounding box parameters). For each $o_m \in \mathcal{O}$, we use RoI-Align~\cite{he2017mask} and average pooling to compute a feature vector $\mathbf{x}^a_{o_m}\in \mathbb{R}^{D_0}$. We also encode the relative locations of conv-features as a spatial feature vector $\mathbf{x}^s_{o_m}$ (See Suppl. for details), which is fused with $\mathbf{x}^a_{o_m}$ to produce the object representation:
\begin{align}
\mathbf{{x}}_{o_m} = F_{vf}([\mathbf{x}^a_{o_m};\mathbf{x}^s_{o_m}])
\end{align}
where $\mathbf{{x}}_{o_m}\in \mathbb{R}^{D}$, $F_{vf}$ is a multilayer network with fully connected layers and $[;]$ is the concatenate operation. 

For the language features, we generate an embedding of noun phrase $p_i \in \mathcal{P}$. To this end, we first encode each word in sentence $Q$ into a sequence of word embedding $\{h_t\}_{t=1\dots T}$ with a Bi-directional GRU~\cite{chung2014empirical}, where $T$ is the number of words in sentence. We then compute the phrase representation $\mathbf{x}_{p_i}$ by taking average pooling on the word representations in each $p_i$:
\begin{align}
\label{phrase}
[h_1, h_2, \dots, h_T] &= \emph{{\rm BiGRU}}_p(Q)\\ 
\mathbf{x}_{p_i} &= \frac{1}{|p_i|}\sum_{t\in p_i} h_t \quad i=1,\cdots,N 
\end{align}
where $\emph{{\rm BiGRU}}_p$ denotes the bi-directional GRU, $h_t,\mathbf{x}_{p_i}\in\mathbb{R}^{D}$ and $h_t= [\stackrel{\rightarrow}{h_t}; \stackrel{\leftarrow}{h_t}]$ is the concatenation of forward and backward hidden states for $t$-th word in the sentence.

\subsection{Phrase Graph Network}

To encode the context dependency among phrases, we now introduce our second module, the phrase graph network, which refines the initial phase embedding features by incorporating phrase relations cues in the description. 

\subsubsection{Phrase Graph Construction}

Specifically, we first build a language scene graph from the image description by adopting an off-the-shelf scene graph parser\footnote{https://github.com/vacancy/SceneGraphParser. We refine the language scene graph for the visual grounding task by rule-based post-processing and more details are included in Suppl.}, which also extracts the phrase relations $\mathcal{R} = \{r_{ij}\}$ from $Q$, where $r_{ij}$ is a relationship phrase that connects $p_i$ and $p_j$.  We denote the language scene graph as $\mathcal{G}_L=\{\mathcal{P}, \mathcal{R} \}$ where $\mathcal{P}$ and $\mathcal{R}$ are the nodes and edges set respectively.
Similar to the phrases in Sec.~\ref{backbone}, we compute an embedding $\mathbf{x}_{r_{ij}}$ for $r_{ij} \in \mathcal{R}$ based on a second bi-directional GRU, denoted as $\emph{{\rm BiGRU}}_r$.

On top of the language scene graph, we construct a phrase graph network that refines the linguistic features through message propagation. Concretely, we associate each node $p_i$ in the graph $\mathcal{G}_L$ with its embedding $\mathbf{x}_{p_i}$, and each edge $r_{ij}$ with its vector representation $\mathbf{x}_{r_{ij}}$. We then define a set of message propagation operators on the graph to generate context-aware representations for all the nodes and edges as follows.    

\subsubsection{Phrase Feature Refinement}\label{phrase_mps}
We introduce two types of message propagation operators to update the node and edge feature respectively. First, to enrich each phrase relation with its subject and object nodes, we send out messages from the noun phrases, which are encoded by their features, to update the relation representation via aggregation:
\begin{align}
\mathbf{x}^c_{r_{ij}} = \mathbf{x}_{r_{ij}} + F^l_{e}([\mathbf{x}_{p_i}; \mathbf{x}_{p_j}; \mathbf{x}_{r_{ij}}])
\end{align}
where $\mathbf{x}^c_{r_{ij}} \in \mathbb{R}^{D}$ is the context-aware relation feature, and $F^l_{e}$ is a multilayer network with fully connected layers.
The second message propagation operator update each phrase node $p_i$ by aggregating features from all its neighbour nodes $\mathcal{N}(i)$ and edges via an attention mechanism: 
\begin{align}
\mathbf{x}^c_{p_i} = \mathbf{x}_{p_i} + \sum_{j\in\mathcal{N}(i)} w_{p_{ij}} F^l_p([\mathbf{x}_{p_j}; \mathbf{x}^c_{r_{ij}}])\label{p-a}
\end{align}
where $\mathbf{x}^c_{p_i}$ is the context-aware phrase feature, $F^l_p$ is a multilayer network, and $w_{p_{ij}}$ is an attention weight between node $p_i$ and $p_j$, which is defined as follows:  
\begin{align}
w_{p_{ij}} = \underset{j\in\mathcal{N}(i)}{\emph{{\rm Softmax}}} (F^l_p([\mathbf{x}_{p_i};\mathbf{x}^c_{r_{ij}}])^\intercal F^l_p([\mathbf{x}_{p_j};\mathbf{x}^c_{r_{ij}}]))
\label{p-w}
\end{align}
Here ${\rm Softmax}$ is a softmax function to compute normalized attention values.

\subsection{Visual Object Graph Network}\label{visgraph}
Similar to the language counterpart, we also introduce a visual scene graph to capture the global scene context for each object proposal, and to build our third module, the visual object graph network, which enriches object features with their contexts via message propagation over the visual graph.

\subsubsection{Visual Scene Graph Construction}
Instead of using a noisy dense graph~\cite{hu2019language}, we propose to construct a visual scene graph relevant to the grounding task by exploiting the knowledge of our phrase graph $\mathcal{G}_L$. To this end, we first prune the object proposal set to keep the objects relevant to the grounding phrases, and then consider only the pairwise relations induced by the phrase graph.

Specifically, we adopt the method in~\cite{flickrentities,GroundingR} to select a small set of high-quality proposals $\mathcal{O}_i$ for each phrase $p_i$. To achieve this, we first compute a similarity score  $\phi^p_{i,m}$ for each phrase-boxes pair $\langle p_i, o_m\rangle$ and a phrase-specific regression offset $\delta^p_{i,m}\in \mathbb{R}^{4}$ for $o_m$ based on the noun phrase embedding $\mathbf{x}^c_{p_i}$ and each object feature $\mathbf{x}_{o_m}$ as follows:
\begin{align}
\phi^p_{i,m} = F_{cls}^p(\mathbf{x}^c_{p_i},\mathbf{x}_{o_m}),\quad
\delta^p_{i,m} = F_{reg}^p(\mathbf{x}^c_{p_i},\mathbf{x}_{o_m})
\end{align}
where $F_{cls}^p$ and $F_{reg}^p$ are two-layer fully-connected networks which transform the input features as in~\cite{mou2016natural}. 

We then select the top $K (K\ll M)$ for each phrase $p_i$ based on the similarity score $\phi^p_{i,m}$, and apply the regression offsets $\delta^p_{i,m}$ to adjust locations of the selected proposals. We denote the refined proposal set of $p_i$ as $\mathcal{O}_i=\{o_{i,k}\}_{k=1}^K$ and all the refined proposals as $\mathcal{V} = \cup_{i=1}^N\mathcal{O}_i$. For each pair of the object proposals $\langle o_{i,k},o_{j,l}\rangle$, we introduce an edge $u_{ij,kl}$ if there is a relation $r_{ij}$ exists in the phrase relation set $\mathcal{R}$. Denoting the edge set as $\mathcal{U} = \{u_{ij,kl}\}$, we define our visual scene graph as $\mathcal{G}_V = \{\mathcal{V},  \mathcal{U}\}$. 

Built on top of the visual scene graph, we introduce a visual object graph network that augments the object features with their context through message propagation. Concretely, as in Sec.~\ref{backbone}, we extract an object feature $\mathbf{{x}}_{o_{i,k}}$ for each proposal $o_{i,k}$ in $\mathcal{V}$. Additionally, for each edge $u_{ij,kl}$ in the graph $\mathcal{G}_V$, we take a union box region of two object $o_{i,k}$ and $o_{j,l}$, which is the minimum box region covering both objects,
and compute its visual relation feature $\mathbf{x}_{u_{ij,kl}}$. To do this, we extract a
convolution feature $\mathbf{x}^a_{u_{ij,kl}}$ from $\mathbf{\Gamma}$ by RoI-Align, and as in the object features, fuse it with a geometric feature $\mathbf{x}^s_{u_{ij,kl}}$ encoding location of two objects (See Suppl. for details). We then develop a set of message propagation operators on the graph to generate context-aware representations for all the nodes and edges in the following.  

\subsubsection{Visual Feature Refinement}
Similar to Sec.~\ref{phrase_mps}, we introduce two types of message propagation operators to refine the object and relation features respectively. Specifically,  we first update relation features by fusing with their subject and object node features:
\begin{align}
\mathbf{{x}}^c_{u_{ij,kl}} = \mathbf{{x}}_{u_{ij,kl}}+ F^v_e([\mathbf{{x}}_{o_{i,k}}; \mathbf{{x}}_{o_{j,l}}; \mathbf{{x}}_{u_{ij,kl}} ])
\end{align}
where $F^v_e$ is a multilayer network with fully connected layers. The second type of message update each object node $o_{i,k}$ by aggregating features from all its neighbour nodes and corresponding edges via the same attention mechanism: 
\begin{align}
\mathbf{ {x} }^c_{o_{i,k}} &= \mathbf{ {x} }_{o_{i,k}} + \sum_{j,l} \alpha_{ij,kl}  F^v_o([\mathbf{ {x} }_{o_{j,l}}; \mathbf{{x}}^c_{u_{ij,kl}}])
\label{v-a}\\
\alpha_{ij,kl}&= \underset{j,l}{\emph{{\rm Softmax}}}(F^v_o([\mathbf{ {x} }_{o_{i,k}}; \mathbf{{x}}^c_{u_{ij,kl}}])^\intercal  F^v_o([\mathbf{ {x} }_{o_{j,l}}; \mathbf{{x}}^c_{u_{ij,kl}}])\nonumber
\end{align}
where $\mathbf{ {x} }^c_{o_{i,k}}$ is the context-aware object feature, $F^v_o$ is a multilayer network and $\alpha_{ij,kl}$ is the attention weight between object $o_{i,k}$ and $o_{j,l}$.
   
\subsection{Graph Similarity Network}
Given the phrase and visual scene graph, we formulate the visual grounding as a graph matching problem between two graphs. To solve this, we introduce a graph similarity network to predict the node and edge similarities between the two graphs, followed by a global inference procedure to predict the matching assignment.

Formally, we introduce a similarity score $\phi_{i,k}$ for each noun phrase and visual object pair $\langle \mathbf{x}^c_{p_i}, \mathbf{ {x} }^c_{o_{i,k}} \rangle$,  and an edge similarity score $\phi_{ij,kl}$ for each phrase and visual relation pair $\langle \mathbf{x}^c_{r_{i,j}}, \mathbf{{x}}^c_{u_{ij,kl}} \rangle$. 
For the \textit{node similarity} $\phi_{i,k}$, we first predict a similarity between the refined features $\langle \mathbf{x}^c_{p_i}, \mathbf{ {x} }^c_{o_{i,k}} \rangle$ as in Sec.~\ref{visgraph}, using two-layer fully-connected networks to compute the similarity score and the object offset as follows,
\begin{align}
\phi^g_{i,k} = F_{cls}^g(\mathbf{x}^c_{p_i} , \mathbf{{x}}^c_{o_{i,k}}) \quad&\quad \delta^g_{i,k} = F_{reg}^g(\mathbf{x}^c_{p_i} , \mathbf{{x}}^c_{o_{i,k}})
\end{align}
We then fuse this with the score used in object pruning to generate the node similarity:  $\phi_{i,k} = \phi^p_{i, k} \cdot \phi^g_{i,k}$. The predicted offset is applied to the proposals in the prediction outcome. 
For the \textit{edge similarity}, we take the same method as in the node similarity prediction, using a multilayer network $F^r_{cls}$ to predict the edge similarity score $\phi_{ij,kl}$:
\begin{align}
\phi_{ij,kl} &= F^r_{cls}(\mathbf{x}^c_{r_{ij}}, \mathbf{{x}}^c_{u_{ij,kl}})
\end{align}

Given the node and edge similarity scores, we now assign each phrase-object pair a binary variable $s_{i,k} \in \{0,1\}$ indicating whether $o_{i,k}$ is the target location of $p_i$. Assuming only one proposal is selected, i.e., $ \sum_{k=1}^K s_{i,k} = 1$, our subgraph matching can be formulated as a structured prediction problem as follows:
\begin{align}
\mathbf{s}^*=& \underset{\mathbf{s}}{\arg\max} 
\big\{ \sum_{i, k} \phi_{i,k} s_{i,k} + \beta\sum_{i,j,k,l} \phi_{ij,kl} s_{i,k}\cdot s_{j,l} \big\} \nonumber\\
s.t.  \; & \sum_{k=1}^K s_{i,k} = 1; \quad i = 1,\dots, N \label{struct-equ}
\end{align}
where $\beta$ is a weight balancing the phrase and relation scores.
We solve the assignment problem by an approximate algorithm based on exhaustive search with a maximal depth~(see Suppl. for detail).

\subsection{Model Learning}
We adopt a pre-trained ResNet-101 network and an off-the-shelf RPN in our backbone network, and train the remaining network modules. 
In order to build the visual scene graph, we adopt a two-stage strategy in our model learning. The first stage learns the phrase graph network and object features by a phrase-object matching loss and a box regression loss. We use the learned sub-modules to select a subset of proposals and construct the rest of our model. The second stage trains the entire deep model jointly with a graph similarity loss and a box regression loss.

Specifically, for a noun phrase $p_i$, the ground-truth for matching scores $\bm{\phi}_i^p= \{ \phi^p_{i, m} \}_{m=1}^M  $ and $\bm{\phi}_i^g=\{ \phi^g_{i,k} \}_{k=1}^K $ are defined as soft label distributions $\bm Y_i^p = \{y_{i,m}^p\}_{m=1}^M $ and $ \bm Y_i^g = \{y_{i,k}^g\}_{k=1}^K$ respectively, based on the IoU between proposal bounding boxes and their ground-truth~\cite{yu2018rethinking}.

Similarly, we compute the ground-truth offset $\delta^{p*}_{i, m}$ between $b_i$ and $o_m$, $\delta^{g*}_{i, k}$ between $b_i$ and $o_{i,k}$.
In addition, the ground-truth for matching scores $\bm{\phi}_{ij}^r=\{\phi_{ij,kl}\}_{k,l=1}^K$  are defined as $ \bm Y_{ij}^r = \{ y_{ij, kl}^r \}_{k,l=1}^K $ based on the IoU between a pair of object proposals $\langle o_{i,k}, o_{j,l} \rangle$ and their ground-truth locations $\langle b_i, b_j \rangle$~\cite{yang2018graph}. 

After normalizing $\bm Y_i^p$, $\bm Y_i^g$ and $\bm Y_{ij}^r$ to probability distributions, we define the matching loss $\mathcal{L}^p_{mat}$ and regression loss $\mathcal{L}^p_{reg} $ in the first stage as follows:
\begin{align}
&\mathcal{L}^p_{mat} = \sum_i  L_{ce}(\bm \phi^p_i, \bm Y_i^p) \nonumber\\
&\mathcal{L}^p_{reg} =\sum_{i}  \frac{1}{||\bm Y_i^p||_0} \sum_{m} \mathbb{I}(y_{i,m}^p>0) L_{sm}(\delta^p_{i,m}, \delta^{p*}_{i, m})
\end{align}
where  $L_{ce}$ is the Cross Entropy loss and $L_{sm}$ is the Smooth-L1 loss. 

For the second stage, the node matching loss $\mathcal{L}^g_{mat}$, edge matching loss $\mathcal{L}^r_{mat}$ and regression loss $\mathcal{L}^g_{reg}$ are defined as:
\begin{align}
\mathcal{L}^g_{mat} &= \sum_i L_{ce}(\bm\phi^g_i, \bm Y_i^g),\quad\mathcal{L}^r_{mat} =\sum_{i,j} L_{ce}(\bm\phi_{ij}^r, \bm Y_{ij}^r) \nonumber  \\
\mathcal{L}^g_{reg} &=\sum_i  \frac{1}{||\bm Y_i^g||_0} \sum_{k}  \mathbb{I}(y_{i,k}^g>0) L_{sm}(\delta^g_{i,k}, \delta^{g*}_{i, k}) 
\end{align}
Here $||*||_0$ is the L0 norm and $\mathbb{I}$ is the indicator function. Finally the total loss $\mathcal{L}$ can be defined as:
\begin{align}
\mathcal{L} &= \mathcal{L}^p_{mat} + \lambda_1\cdot \mathcal{L}^p_{reg}    
\nonumber\\
&+ \lambda_2\cdot \mathcal{L}^g_{mat} + \lambda_3\cdot  \mathcal{L}^r_{mat} + \lambda_4\cdot \mathcal{L}^g_{reg}
\end{align}
where $\lambda_1$, $\lambda_2$, $\lambda_3$, $\lambda_4$ are weighting coefficients for balancing loss terms.


\section{Experiments}
\subsection{Datasets and Metrics}
We evaluate our approach on Flickr30K Entities~\cite{flickrentities} dataset, which contains 32k images, 275k bounding boxes, and 360k noun phrases. Each image is associated with five sentences description and the noun phrases are provided with their corresponding bounding boxes in the image. Following~\cite{GroundingR}, if a single noun phrase corresponds to multiple ground-truth bounding boxes, we merge the boxes and use the union region as their ground-truth. We adopt the standard dataset split as in Plummer et al.~\shortcite{flickrentities}, which separates the dataset into 30k images for training, 1k for validation and 1k for testing. 
We consider a noun phrase grounded correctly when its predicted box has at least 0.5 IoU with its ground-truth location. The grounding accuracy (i.e., Recall@1) is the fraction of correctly grounded noun phrases.

\subsection{Implementation Details}
We generate an initial set of $M=100$ object proposals with a RPN from Anderson et al.~\shortcite{anderson2018bottom}\footnote{It is based on FasterRCNN~\cite{ren2015faster} with ResNet-101 as its backbone, trained on Visual Genome dataset~\cite{krishna2017visual}. We use its RPN to generate object proposals.}. We use the output of ResNet C4 block as our feature map $\mathbf{\Gamma}$ with channel dimension $D_0=2048$ and the visual object features are obtained by applying RoI-Align with resolution $14\times 14 $ on $\mathbf{\Gamma}$. The embedding dimension $D$ of phrase and visual representation is set as $1024$. In visual graph construction, we select the most $K=10$ relevant object candidates for each noun phrase.

For model training, we use SGD optimizer with initial learning rate 5e-2, weight decay 1e-4 and momentum 0.9. We train 60k iterations with batch-size 24 totally and decay the learning rate 10 times in 20k and 40k iterations respectively. 
The loss weights of regression terms $\lambda_1$ and $\lambda_4$ are set to 0.1 while matching terms $\lambda_2$ and $\lambda_3$ are set to 1. 
During the test stage, we search an optimal weight $\beta^*\in [0,1]$ on val set and apply it to test set directly.

\subsection{Results and Comparisons}

\begin{table}
	\centering
	\caption{\small {Results Comparison on Flickr30k test set.}}
	\resizebox{0.4\textwidth}{!}{
		\begin{tabular}{cccc}
			\hline
			Methods                                &\textbf{Accuracy(\%)} \\ 
			\hline
		
			SMPL\cite{wang2016structured}                        &42.08  \\
			NonlinearSP~\cite{wang2016learning}                       &43.89  \\
			GroundeR~\cite{GroundingR}                              &47.81  \\
			MCB~\cite{fukui2016multimodal}                           &48.69  \\
			RtP~\cite{flickrentities}                                &50.89  \\
			Similarity Network~\cite{wang2018learning}                &51.05  \\
			IGOP~\cite{yeh2017interpretable}                          &53.97  \\
			SPC+PPC \cite{plummerPLCLC2017}                          &55.49  \\
			SS+QRN \cite{chen2017query}                               &55.99  \\
			CITE \cite{plummer2018conditional}                        &59.27  \\
            SeqGROUND \cite{SeqGROUND}                         &61.60  \\
            \textbf{Our approach~(ResNet-50)}                 &\textbf{67.90} \\ 
            \hline
            \hline
            DDPN \cite{yu2018rethinking}                              &73.30 \\
			\textbf{Our approach~(ResNet-101)}                &\textbf{76.74} \\ 
			\hline
	\end{tabular}}
	\label{flickr-results}
\end{table}

\begin{table}
	\centering
	
	\caption{\small{Comparison of phrases grounding accuracy over coarse categories on Flickr30K test set.}}
	\resizebox{0.475\textwidth}{!}{
		\begin{tabular}{ccccccccc}
			\hline
			Methods         &people&clothing&bodyparts&animal&vehicles&instruments&scene&other  \\ 
            \hline
			SMPL            &57.89 &34.61 &15.87 &55.98 &52.25 &23.46 &34.22 &26.23 \\
			GroundR         &61.00 &38.12 &10.33 &62.55 &68.75 &36.42 &58.18 &29.08 \\
			RtP             &64.73 &46.88 &17.21 &65.83 &68.72 &37.65 &51.39 &31.77 \\
            IGOP            &68.17 &56.83 &19.50 &70.07 &73.72 &39.50 &60.38 &32.45 \\
            SS+QRN          &68.24 &47.98 &20.11 &73.94 &73.66 &29.34 &66.00 & 38.32 \\
            SPC+PPC         &71.69 &50.95 &25.24 &76.23 &66.50 &35.80 &51.51 &35.98 \\  
			CITE            &73.20 &52.34 &\textbf{30.59} &76.25 &75.75 &48.15 &55.64 &42.83 \\
			SeqGROUND       &76.02 &56.94 &26.18 &75.56 &66.00 &39.36 &\textbf{68.69} &40.60 \\
            \hline
            \hline
			\textbf{Ours~(RN-50)} &\textbf{83.06}&\textbf{63.35}&24.28&\textbf{84.94}&\textbf{78.25}&\textbf{55.56}&61.67&\textbf{52.05} \\
			\textbf{Ours~(RN-101)}&\textbf{86.82}&\textbf{79.92}&\textbf{53.54}&\textbf{90.73}&\textbf{84.75}&\textbf{63.58}&\textbf{77.12}&\textbf{58.65}
			\\
			\hline
	\end{tabular}}
	\label{flickr-aba-detail} 
\end{table}

We report the performance of the proposed framework on the Flickr30K Entities test set and compare it with several the state-of-the-art approaches. Here we consider two model configurations for proper comparisons, which use an ResNet-50\footnote{Model details of ResNet-50 backbone are included in Suppl.} and an ResNet-101 as their backbone network, respectively.
 
As shown in Tab.~\ref{flickr-results}, our approach outperforms the prior methods by a large margin in both settings. In particular, our model with ResNet-101 backbone achieves \textbf{76.74\%} in accuracy, which improves 3.44\% compared to DDPN~\cite{yu2018rethinking}.
For the setting that uses ResNet-50 backbone and a pretrained RPN on MSCOCO~\cite{MSCOCO} dataset,
we can see that our model achieves \textbf{67.90\%} in accuracy and outperforms SeqGROUND by 6.3\%. We also show detailed comparisons per coarse categories in Tab.~\ref{flickr-aba-detail} and it is evident that our approach achieves better performances consistently on most categories. 
\begin{table*}[t]
	\begin{minipage}{.7\textwidth}
		\centering
		\caption{\label{flickr-aba-val}\small{Ablation study on Flickr30K val set.}}
		\resizebox{0.9\textwidth}{!}{
			\begin{tabular}{cccccc|cccc}
				\hline
				& \multicolumn{4}{c}{\textbf{Components}}& &   \multicolumn{4}{c}{\textbf{Components (w/o relations feature)}} \\
				Methods &PGN&PP&VOGN&SP&\textbf{Acc(\%)}&PGN&PP&VOGN&\textbf{Acc(\%)} \\
				
				\hline
				Baseline &-&-&-&-                                    &73.46 
				&-&-&-                                               &-\\
				
				&\checkmark&-&-&-                                    &74.40 
				&\checkmark\small{(w/o $\mathbf{x}^c_{r_{ij}}$)}&-& - &74.11\\
				
				&\checkmark&\checkmark&-&-                           &75.50 
				&\checkmark\small{(w/o $\mathbf{x}^c_{r_{ij}}$)}&\checkmark&-&75.32\\	
				&\checkmark&\checkmark&\checkmark&-                  &75.85 
				&\checkmark\small{(w/o $\mathbf{x}^c_{r_{ij}}$)}&\checkmark&\checkmark\small{(w/o $\mathbf{x}_{u_{ij,kl}}^c$)}&75.44\\
				Ours&\checkmark&\checkmark&\checkmark&\checkmark&\textbf{76.19}
				&-&-&-&-\\
				\hline
			\end{tabular}
		}
		
	\end{minipage}
	\begin{minipage}{.1\textwidth}
		\centering
		\parbox{5cm}{\caption{\label{flickr-aba-k}\small{Ablation study of $K$ proposals on Flickr30K val set.}}}
		\resizebox{2.7\textwidth}{!}{
			\begin{tabular}{c|c|c|c}
				\hline 
				$K$ &  5 & 10 & 20 \\ 
				\hline
				Acc(\%) & 74.97 & \textbf{76.19} & 76.07\\
			\end{tabular}
		}
	\end{minipage}
\end{table*}

\begin{figure*}[t]
	\centering
	\includegraphics[width=0.85\linewidth]{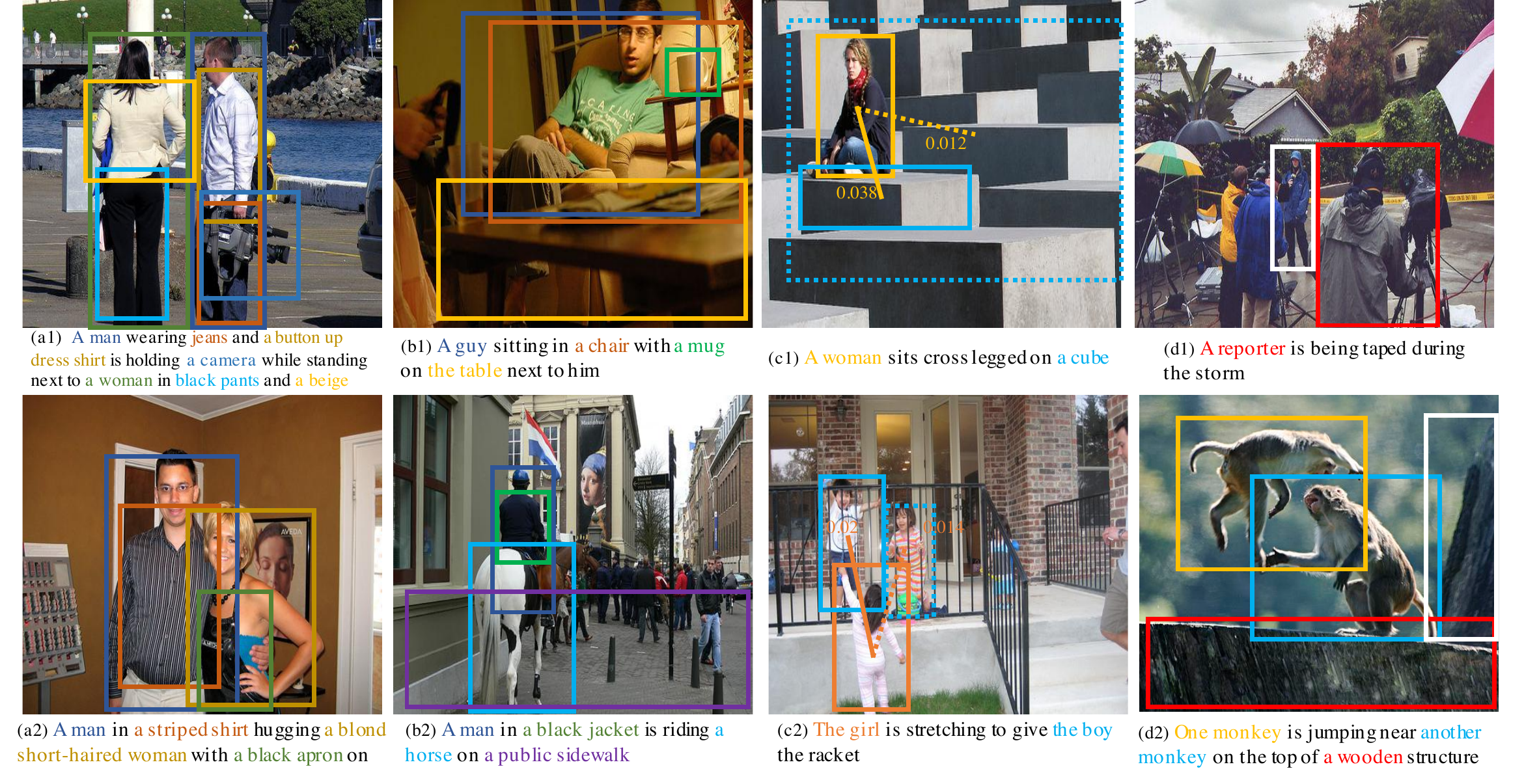}
	\caption{\small{Visualization of phrase grounding results in Flickr30K val set. The colored bounding boxes, which are predicted by our approach, correspond to the noun phrases in the sentences with the same color. The dot boxes denote the predicted results without relations constraint, while the white boxes are ground-truths and the red boxes are the incorrect predictions. The last column is the failure cases.}}
	\label{fig}
\end{figure*}

\subsection{Ablation Studies}


In this section, we perform several experiments to evaluate the effectiveness of individual components, investigate hyper-parameter $K$ and the impact of relations feature in two graphs in our framework with ResNet-101 as the backbone on Flickr30k val set\footnote{We include ablations of ResNet-50 backbone in Suppl.}, which is shown in Tab.~\ref{flickr-aba-val} and Tab.~\ref{flickr-aba-k}.

\subsubsection{Baseline:}
The baseline first predicts the similarity score and regression offset for each phrase-box pair $\langle \mathbf{x}_{p_i}, \mathbf{x}_{o_m} \rangle$, and then selects the most relevant proposal followed by applying its offset. Our baseline grounding accuracy achieves 73.46\% with ResNet-101 backbone.

\subsubsection{Phrase Graph Net~(PGN):}PGN propagate language context cues via the scene graph structure effectively. The noun phrases feature can not only be aware of long-term semantic contexts from the other phrases but also enriched by its relation phrases representation. The experiment shows that our PGN can improve the accuracy from 73.46\% to 74.40\%.

\subsubsection{Proposal Pruning~(PP):} 
The quality of proposals generation plays an important role in visual grounding task. Here we take proposal pruning operation by utilizing PGN, which can help reduce more ambiguous object candidates with language contexts. We can see a significant improvement of $1.1\%$ accuracy.

\subsubsection{Visual Object Graph Net~(VOGN):}
When integrating the VOGN into the whole framework, we can achieve $75.85\%$ accuracy, which is better than the direct matching with the phrase graph. This suggests that the object representation can be more discriminative after conducting message passing among context visual object features\footnote{See Suppl. for more experiments that analyze the VOGN.}.

\subsubsection{Structured Prediction~(SP):} The aforementioned PGN and VOGN take the context cues into consideration during their nodes matching. Our approach, by contrast, explicitly takes the cross-modal relation matching into account and predicts the final result via a global optimization. We can see further improvement of accuracy from 75.85\% to 76.19\%.

\subsubsection{Hyper-parameter $K$ and Relations Feature:}
In Tab.\ref{flickr-aba-k}, our framework achieves the highest accuracy when $K=10$ while $K=5$ will result in performance dropping from 76.19\% to 74.97\% due to the lower proposals recall. When $K=20$, our model will get a comparable performance but consume more computation resources and inference time.

We also perform experiments to show the impact of relation phrases and visual relations in PGN and VOGN in Tab.~\ref{flickr-aba-val}.
For PGN, the performance will drop from 74.40\% to 74.11\% without phrase relations  $\mathbf{x}_{r_{ij}}^c$. 
And we can see 0.41\% performance drop when ignoring both phrase relations $\mathbf{x}_{r_{ij}}^c$ and visual relations $\mathbf{x}_{u_{ij,kl}}^c$ in PGN and VOGN.

\subsection{Qualitative Visualization Results}
We show some qualitative visual grounding results in Fig.\ref{fig} to demonstrate the capabilities of our framework in challenging scenarios.
In (a1) and (a2), our framework is able to successfully localize multiple entities in the long sentences without ambiguity.
With the help of VOGN, we can see that our model localize \textit{a mug} close to man correctly rather than another mug in the left bottom in (b1). Column 3 shows that relations constraint can help refine the final prediction. 
The last column is failure cases. Our model cannot ground objects in images correctly with severe visual ambiguity.

\section{Conclusion}
In this paper, we have proposed a context-aware cross-modal graph network for visual grounding task. 
Our method exploits a graph representation for language description, and transfers the linguistic structure to object proposals to build a visual scene graph. Then we use message propagation to extract global context representations both for the grounding entities and visual objects. As a result, it is able to conduct a global matching between both graph nodes and relation edges.
We present a modular graph network to instantiate our core idea of context-aware cross-modal matching. 
Moreover, we adopt a two-stage strategy in our model learning, of which the first stage learns a phrase graph network and visual object features while the second stage trains the entire deep network jointly.
Finally, we achieve the state-of-the-art performances on Flickr30K Entities benchmark, and outperform other approaches by a sizable margin.

{\small
	\bibliographystyle{aaai}
	\bibliography{egbib}
}
\end{document}


\maketitle
\section{Cross-modal Graph Network}

\subsection{Spatial Feature of Object}
We generate a coordinate map $\mathbf{\alpha}$ with the same spatial size as the convolution feature map $\mathbf{\Gamma}$. The coordinate map $\mathbf{\alpha}$ consists of two channels, indicating the x, y coordinates for each pixel in $\mathbf{\Gamma}$, and normalized by the feature map center. For each object proposal $o_m\in \mathbb{R}^4$, we crop a coordinate map from $\mathbf{\alpha}$ with RoI-Align and embed it into a spatial feature vector $\mathbf{x}^s_{o_m}\in\mathbb{R}^{256}$ by multiple fully connection layers. 
\begin{figure}[ht]
	\centering
	\includegraphics[width=0.7\linewidth]{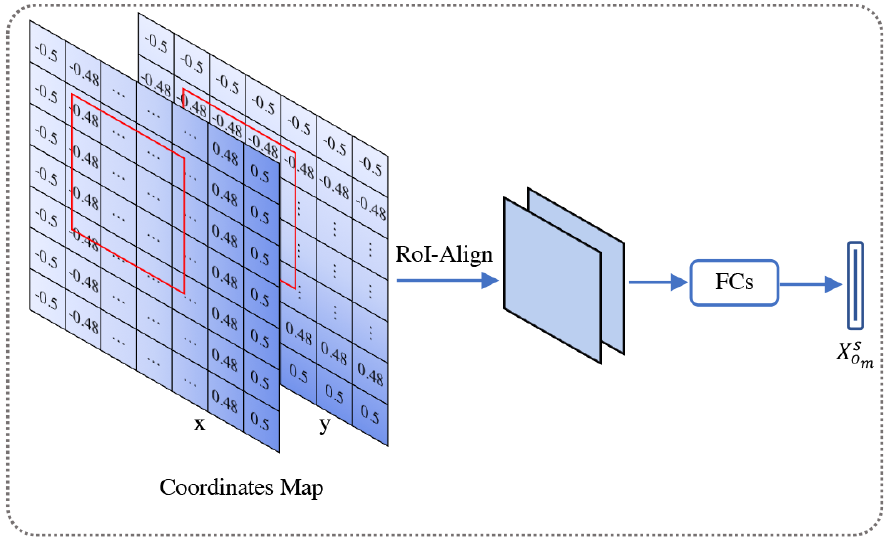}
	\caption{\small{Illustration of spatial feature embedding.}}
\end{figure}

\subsection{Spatial Feature of Union Region}
We generate a two-channel binary mask for $o_{i,k}$ and $o_{j,l}$ separately where locations within object proposal $o_{i,k}$ , $o_{j,l}$ fill 1 and others fill 0. Then the two-channel binary mask is resized to $64\times64$. And we use multiple fully connected layers to embed it to a geometric feature vector $\mathbf{x}^s_{u_{ij,kl}}\in \mathbb{R}^{256}$.

\begin{figure}[ht]
	\centering
	\includegraphics[width=0.7\linewidth]{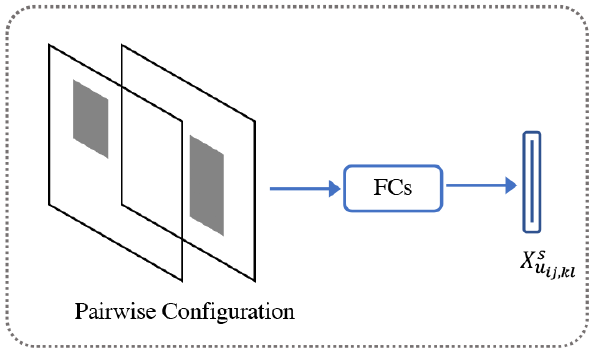}
	\caption{\small{Illustration of pairwise geometric feature embedding.}}
\end{figure}

\subsection{Scene Graph Parser}
For a given sentence, we use a public toolkit\footnote{https://github.com/vacancy/SceneGraphParser} to generate a language scene graph, in which nodes encode noun phrases and edges are the relationships between them. In this language scene graph parser, a dependency parser is first applied to the input sentence and then hand-crafted rules are employed to generate language scene graphs. However, we observe some issues associated with the off-the-shelf parser: 1) noun phrases in the parses sometimes do not correspond to the given phrases; 2) some phrases and their relationships are still missing the in parses.

To address the aforementioned limitations, we perform additional post-processing on the Flickr30K Entities dataset. First, we take all given phrases as graph nodes. For each phrase, we pick a noun phrase in the parse that has a maximum word overlap with this given phrase. We then assign the parsed relations to these nodes. However, there are still some 
isolated nodes in the resulting graph. We further recall some missing relations by taking advantage of the coarse categories of the given phrases. Specifically, for an isolated phrase, if its type is \textit{clothing} or \textit{bodyparts},  we find a phrase with the type of \textit{people} as its subject, and assign a relationship \textit{wear} / \textit{have} to them. If there are multiple phrases with the type of \textit{people} in the graph nodes, we select the one that has a minimum word distance in the sentence with the isolated phrase. The motivation of our rules design comes from the observation that most of \textit{clothing} / \textit{bodyparts} phrases are related to a \textit{people} phrase, and their relationships are generally \textit{wear} / \textit{have}.

\subsection{Solving Structured Prediction}
We solve the structured prediction problem by taking an exhaustive search on all the possibilities of $\bm s$ in Equ. 12 with a maximal depth when noun phrase number $N$  is less than 6, and applying only node matching between the phrase graph and visual scene graph otherwise. The motivation of the solving strategy comes from the observation that 96.12\% language scene graphs in Flickr30K dataset have less than 6 nodes. The complexity of exhaustive search with a maximal depth is $K^N$, which is not time-consuming when $N$ is small. 

\section{Experiments}
\subsection{Model details with ResNet-50 backbone}
We take an off-the-shelf object detector with ResNet-50 as its backbone to generate the initial set of proposals. It is based on FasterRCNN and pre-trained on the MSCOCO dataset~(Lin et al.2014). Other settings are same to the model with ResNet-10 backbone. During the training stage, we use SGD optimizer with initial learning rate 1e-1, weight decay 1e-4 and momentum 0.9. The model is trained with 60k iterations totally with batch size 24, and decay the learning rate 10 times in 20k and 40k iterations respectively.

\subsection{Ablations with ResNet-50 backbone}

\begin{table}[ht]
	
	\centering
	\caption{\small{Ablation studies on Flickr30K val set with ResNet-50 backbone.}}
	\resizebox{0.35\textwidth}{!}{
		\begin{tabular}{cccccc}
			\hline
			& \multicolumn{4}{c}{\textbf{Components}}&   \\
			Methods &PGN&PP&VOGN&SP&\textbf{Acc(\%)}  \\ 
			
			\hline
			\hline
			
			Baseline    &-&-&-&-                                 &60.31 \\
			&\checkmark&-&-&-                                    &62.45  \\
			&\checkmark&\checkmark&-&-                           &67.51  \\
			&\checkmark&\checkmark&\checkmark&-                  &67.77  \\
			Ours  &\checkmark&\checkmark&\checkmark&\checkmark   &68.12  \\
			\hline
	\end{tabular}}
	\label{comp-aba-val} 
\end{table}

In order to investigate the effectiveness the individual component of our framework with ResNet-50 backbone, we also conduct a series of ablation studies. As shown in Tab.~\ref{comp-aba-val}, the accuracy shows the same growth trend compared to ResNet-101 backbone. In particular, we can observe a significant performance improvement when adopting proposal pruning over baseline model, which improves the accuracy from 60.31\% to 66.77\%. 
This indicates that proposal pruning is critical for visual grounding task when the object detector doesn't perform well.

	
			
			

\subsection{Additional Experiments on VOGN}
\begin{table}
	\centering
	\caption{\small{Additional Experments of VOGN with ResNet-101 backbone on Flickr30K val set}}
	\resizebox{0.4\textwidth}{!}{
		\begin{tabular}{cccccc}
			\hline
			& \multicolumn{4}{c}{\textbf{Components}}&   \\
			Methods &PGN&PP&VOGN&SP&\textbf{Acc(\%)}  \\ 
			
			\hline
			\hline
			
			Baseline    &-&-&-&-                                            &73.46 \\
			&-&\checkmark&-&-                                            &74.60  \\
			&-&\checkmark&\checkmark&-                                            &75.59  \\
			&-&\checkmark&\checkmark(w/o $x^c_{u_{ij,kl}}$)&-            &74.80  \\
			\hline
	\end{tabular}}
	\label{vogn} 
\end{table}
To validate the effectiveness of VOGN, we conduct some additional experiments as shown in Tab.~\ref{vogn}.
In the baseline model, we compute the similarity score and regression offset for each phrase-box pair$\langle \mathbf{x}_{p_i}, \mathbf{x}_{o_m} \rangle$. Then we adopt proposal pruning strategy over baseline model without PGN, which can improve grounding accuracy from 73.46\% to 74.6\%.  Furthermore, we add our VOGN under this setting and observe a significant improvement from 74.60\% to 75.59\%, which indicates the visual object representation can be more discriminative with its context cues. 

Finally, the performance will drop sharply from 75.59\% to 74.80\%  without considering visual relations feature $\mathbf{x}^c_{u_{ij,kl}}$ during message passing, which suggests that visual relations play an important role in computing attention among objects.